\documentclass[runningheads]{llncs}

\usepackage[T1]{fontenc}

\usepackage{graphicx}

\usepackage{booktabs}
\usepackage[misc]{ifsym}

\usepackage{mwe}

\usepackage{footnote}
\usepackage{footnotebackref}
\usepackage{multirow}
\usepackage{makecell}
\usepackage{algorithm}
\usepackage{algorithmic}
\usepackage{amsmath}

\usepackage{float}
\usepackage{subfigure}
\usepackage{tikz}
\usetikzlibrary{shapes,arrows}
\usepackage{hyperref}
\usepackage{appendix}
\usepackage[skins]{tcolorbox}
\tcbuselibrary{breakable}
\usepackage{bm}
\usepackage{amsbsy}
\usepackage{amsfonts}
\usepackage{etoolbox}

\begin{document}

\title{Subgraph Retrieval Enhanced by Graph-Text Alignment for Commonsense Question Answering}

\titlerunning{SEPTA}

\author{Boci Peng\inst{1} \and
Yongchao Liu\inst{2} \and
Xiaohe Bo\inst{3} \and
Sheng Tian\inst{2} \and
Baokun Wang\inst{2} \and
Chuntao Hong\inst{2} \and
Yan Zhang\inst{1}\textsuperscript{(\Letter)}}
\authorrunning{Peng et al.}

\tocauthor{Boci Peng, Yongchao Liu, Xiaohe Bo, Sheng Tian, Baokun Wang, Chuntao Hong, Yan Zhang}
\toctitle{Subgraph Retrieval Enhanced by Graph-Text Alignment for Commonsense Question Answering}

\institute{School of Intelligence Science and
Technology, Peking University, China\\
\email{bcpeng@stu.pku.edu.cn}, \email{zhyzhy001@pku.edu.cn}\and Ant Group, China\\
\email{\{yongchao.ly,tiansheng.ts,yike.wbk,chuntao.hct\}@antgroup.com}\and
School of Artificial Intelligence, Beijing Normal University, China\\
\email{xiaohe@mail.bnu.edu.cn}}

\maketitle

\begin{abstract}
Commonsense question answering is a crucial task that requires machines to employ reasoning according to commonsense. Previous studies predominantly employ an extracting-and-modeling paradigm to harness the information in KG, which first extracts relevant subgraphs based on pre-defined rules and then proceeds to design various strategies aiming to improve the representations and fusion of the extracted structural knowledge. Despite their effectiveness, there are still two challenges. On one hand, subgraphs extracted by rule-based methods may have the potential to overlook critical nodes and result in uncontrollable subgraph size. On the other hand, the misalignment between graph and text modalities undermines the effectiveness of knowledge fusion, ultimately impacting the task performance. To deal with the problems above, we propose a novel framework: \textbf{S}ubgraph R\textbf{E}trieval Enhanced by Gra\textbf{P}h-\textbf{T}ext \textbf{A}lignment, named \textbf{SEPTA}. Firstly, we transform the knowledge graph into a database of subgraph vectors and propose a BFS-style subgraph sampling strategy to avoid information loss, leveraging the analogy between BFS and the message-passing mechanism. In addition, we propose a bidirectional contrastive learning approach for graph-text alignment, which effectively enhances both subgraph retrieval and knowledge fusion. Finally, all the retrieved information is combined for reasoning in the prediction module. Extensive experiments on five datasets demonstrate the effectiveness and robustness of our framework.

\end{abstract}

\section{Introduction}

Commonsense question answering (CSQA) is a critical task in natural language understanding, which requires systems to acquire different types of commonsense knowledge and possess multi-hop reasoning ability~\cite{ref:csqa,ref:obqa,ref:socialiqa}. Though massive pre-trained models have achieved impressive performance on this task, it is difficult to learn commonsense knowledge solely from the pre-training text corpus, as the commonsense knowledge is evident to humans and rarely expressed explicitly in natural language. Compared with unstructured text, structured data like knowledge graphs is much more efficient in representing commonsense~\cite{ref:jointlk}. The incorporation of external knowledge aids PLMs in comprehending question-answer (Q-A) pairs, while the entity relations enhance the model’s reasoning capabilities. Therefore, various commonsense knowledge graphs (CSKGs) (e.g., ConceptNet~\cite{ref:cpnet}) have been adopted in previous studies.

Existing KG-augmented models for CSQA primarily adhere to a extracting-and-modeling paradigm~\cite{ref:greaselm,ref:jointlk,ref:gsc,ref:dhlk,ref:grape,ref:fits}. First, the knowledge subgraphs or paths related to a given question are extracted by string matching or semantic similarity, which indicate the relations between concepts or imply the process of multi-hop reasoning. Subsequently, diverse strategies emerge for the efficient representation and fusion of the extracted structural knowledge. One research path~\cite{ref:safe,ref:hamqa} involves elaborately crafting graph neural networks for better modeling the extracted subgraphs, whereas another~\cite{ref:qagnn,ref:jointlk} explores the efficient incorporation of knowledge from KG into language models by enhancing the interactions between PLMs and GNNs.

Despite their success, these approaches still have several limitations. First, the subgraph's quality suffers when retrieved through a simple string or semantic matching, posing limitations for subsequent operations. To obtain sufficient relevant knowledge, the number of nodes will expand dramatically with the increase of hop count, inevitably raising the burden of the model. Despite its ample size, certain crucial nodes might remain elusive, since some entities are not learned during the pre-training. Besides, the edges linked to the peripheral nodes within the subgraph are pruned, causing the message-passing mechanism of GNN to be blocked and impairing the attainment of effective representations, consequently undermining valuable information. Second, the misalignment between graph and text encoders presents a challenge for PLMs to internalize the knowledge contained in the acquired subgraph, especially in scenarios with limited data, leading to a reduced task performance~\cite{ref:fits}. Though Dragon~\cite{ref:dragon} proposes a pre-training method to align GNNs and PLMs, it requires additional corpus, and the text-to-graph style to construct semantically equivalent graph-text pairs is challenging. The necessity for substantial computational resources poses another hurdle, prompting the search for a more efficient alignment method.

In this paper, we propose a novel framework: \textbf{S}ubgraph R\textbf{E}trieval Enhanced by Gra\textbf{P}h-\textbf{T}ext \textbf{A}lignment (\textbf{SEPTA}), for CSQA. To mitigate the shortcomings of the subgraph extraction process, we establish a database of subgraph vectors derived from the knowledge graph. Consequently, the challenge shifts from retrieving a pertinent subgraph to obtaining relevant subgraph vectors. A BFS-style sampling method is employed to obtain the connected graph for each node and the embedding of the subgraph is subsequently stored in the database. Drawing on the parallels between BFS and the message-passing mechanism of GNNs, the central node's representation learned from the subgraph could be closely aligned with that derived from the entire graph, with almost no information loss. Besides, to further improve the retrieval accuracy and facilitate knowledge fusion during the prediction, we consider aligning the semantic space of the graph and text encoders, proposing an effective approach for graph-text alignment. A novel graph-to-text method is proposed to construct high-quality semantically equivalent training pairs, with no requirement of external corpus and easy to train. Finally, all the information retrieved is combined by a simple attention mechanism to facilitate the model in commonsense reasoning.

Our contributions can be summarized as follows:
\begin{itemize}
\item We propose a novel and effective framework SEPTA, where we convert the knowledge graph into a subgraph vector database and retrieve relevant subgraphs to facilitate commonsense reasoning.

\item We design a bidirectional contrastive learning method to align the semantic space of the graph and text encoders, with a graph-to-text method to construct high-quality graph-text pairs, which facilitates subgraph retrieval and knowledge fusion.

\item We propose a BFS-style subgraph sampling strategy for subgraph construction. Drawing on the parallel between BFS and the message-passing mechanism, our method can preserve complete neighbor information for each node.

\item We conduct extensive experiments on five datasets. Our proposed approach achieves better results than the state-of-the-art approaches and has promising performance in weakly supervised settings.

\end{itemize}

\section{Related Work}
\subsection{Commonsense Question Answering}
Commonsense question answering aims to evaluate the reasoning ability of models based on commonsense knowledge~\cite{ref:commonsenseqa}, e.g., physical commonsense~\cite{ref:piqa}. To incorporate external knowledge and enhance reasoning ability, some works introduce commonsense knowledge graphs (CSKGs, e.g. ConceptNet~\cite{ref:cpnet}). Generally, these methods~\cite{ref:qagnn,ref:dgrn,ref:greaselm,ref:jointlk,ref:gsc,ref:safe,ref:dragon,ref:hamqa,ref:dhlk,ref:grape,ref:fits} extract relevant knowledge subgraphs through entity linking and adopt graph neural networks to learn knowledge representations. Among them, a category of research focuses on designing more efficient knowledge encoders. For example, SAFE~\cite{ref:safe} proposes a 2-layer MLP to improve the efficiency of graph encoding. HamQA~\cite{ref:hamqa} considers learning hierarchical structures in KGs with hyperbolic geometry. Another research line tries to enhance the interactions between PLMs and GNNs. For instance, QA-GNN~\cite{ref:qagnn} adds a QA context node to the retrieved subgraphs and incorporates relevant information from other entities. Unlike previous works, we convert the knowledge graph into a subgraph database and transform the task to a subgraph vector retrieval problem, thus bypassing the challenges inherent in the extracting-and-modeling paradigm. 

\subsection{Graph-Text Alignment}
Aligning the embedding spaces of text encoders and graph encoders is an effective way to take the strengths of two modalities~\cite{ref:survey}. Previous alignment methods can be roughly classified into two groups, i.e. the symmetric method and the asymmetric method, based on their training objectives. The symmetric alignments enhance each modality equally, most of which adopt a two-tower style and utilize contrastive learning techniques~\cite{ref:safer,ref:congrat}. However, the asymmetric methods aim to take advantage of the capabilities of GNNs to reinforce PLMs. The predominant approaches can be categorized into two types, with the first type trying to enhance PLMs by inserting graph encoders into transformers~\cite{ref:patton,ref:yunzhu} and the second type directly using GNNs as teacher models to generate soft labels for the PLMs~\cite{ref:Heterogeneous}. In this paper, we leverage PLM as a teacher model and distill its knowledge into GNN, enabling us to retrieve related subgraphs in a manner akin to retrieving relevant text. Although DRAGON~\cite{ref:dragon} also proposes a pre-training method to align PLMs and GNNs, our approach does not require additional corpus and demands lower computational costs.

\section{Task Formulation}
We study the multiple-choice CSQA~\cite{ref:csqa,ref:obqa}, which can be formulated as: given a natural language question $q$ and a set of answer candidates $C = \{c_1, c_2, \dots, c_n\}$, the aim is to identify the optimal choice $c^* \in C$. Consistent with previous works~\cite{ref:kagnet}, the CSQA problem is addressed in a \emph{knowledge-aware} setting, that is, we can utilize external commonsense knowledge graphs (CSKGs) to facilitate model prediction. A CSKG can be formally described as a multi-relational graph $\mathcal{G} = (\mathcal{V}, \mathcal{R}, \mathcal{E}, \boldsymbol{X})$, where $\mathcal{V}$ is the set of concept nodes (e.g., \emph{Sun} and \emph{Holiday}), $\mathcal{R}$ is the set of relation types (e.g., \emph{HasProperty} and \emph{AtLocation}), $\mathcal{E} \in \mathcal{V} \times \mathcal{E} \times \mathcal{V}$ is the link set of the knowledge graph (or fact triplets, e.g. \emph{(House, MadeOf, Wood)}), and $\boldsymbol{X} \in \mathbb{R}^{\lvert \mathcal{V} \rvert \times d}$ denotes pre-trained embedddings of all concept nodes. Generally, the task can be treated as a score prediction task for each Q-A pair. 

\section{Methods}

\begin{figure*}[t!]
    \centering
    \setlength{\abovecaptionskip}{0.15cm}
    \includegraphics[scale=0.39]{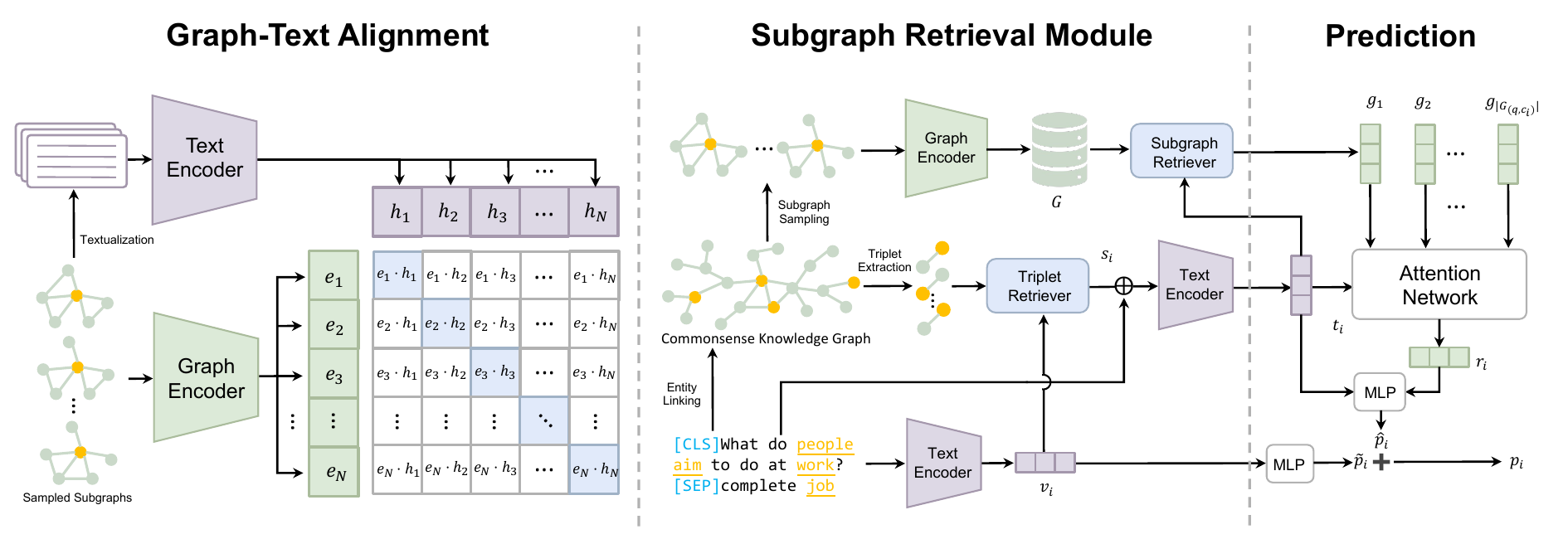}
    \caption{The overview of our proposed SEPTA. First, a bidirectional contrastive method is proposed to align the semantic space of graph and text encoders. With the encoders aligned, we then transform the knowledge graph into a subgraph vector database and introduce a query enhancement strategy for better subgraph retrieval. Finally, all the information retrieved is combined by a simple attention mechanism to bolster the reasoning ability of PLMs for CSQA.}
    \label{fig:overview}
\end{figure*}

In this section, we will introduce the design of our SEPTA. Departing from previous extracting-and-modeling approaches, we reframe the task as a subgraph vector retrieval problem and propose a graph-text alignment method to improve the retrieval accuracy and facilitate knowledge fusion for prediction. For ease of exposition, we first introduce the graph-text alignment process in Section~\ref{sec:alignment}. Then with the aligned encoder, the subgraph vector database is constructed and retrieved, which will be presented in Section~\ref{sec:retrieval}. Finally, how to combine all the structural information retrieved for answer prediction is discussed in Section~\ref{sec:prediction}. Figure~\ref{fig:overview} shows the overview of our SEPTA.

\subsection{Graph-Text Alignment}
\label{sec:alignment}
To coordinate the embedding spaces of graph and text encoders and fully harness the respective strengths of text and KG, we propose an alignment process before downstream tasks. In our method, we initially address the challenge of generating training graph-text pairs with equivalent semantics and subsequently employ a bidirectional contrastive learning method to train the encoders of both modalities. The alignment process plays a pivotal role, as for one thing, it decides the efficacy of retrieving question-related subgraphs from the vector base, and for another, it determines the successful integration of graph information with the question context during the prediction phase.

\subsubsection{Construction of Graph-Text Pairs}
The construction of high-quality semantically equivalent graph-text pairs is crucial for the alignment process, yet not that straightforward. Previous methods mostly adopt a text-to-graph approach to construct training pairs, where the goal is to discover a graph structure that corresponds to the semantics of a provided text segment. However, utilizing existing transformation tools e.g. dependency graph could not well accommodate the downstream subgraph retrieval, while rule-based methods to extract text-related subgraphs from CSKGs are challenging. It is also notably time-consuming and laborious to construct through manual annotation. Therefore, in this paper, we propose a graph-to-text approach and consider constructing synonymous text descriptions of the subgraphs.

Specifically, we propose a BFS-style sampling strategy for subgraph construction, which initiates from the central node and proceeds to sample neighbors layer by layer. During the process, to address the challenge of an excessive number of neighbors, we set $p$ as the probability for immediate neighbor selection. In addition, since it is sufficient to describe local neighborhoods for determining structural equivalence~\cite{ref:node2vec}, we set a parameter $d$ to constrain the depth of the sampled subgraphs. By restricting the search to nearby nodes, our sampling method achieves this characterization and obtains a microscopic view of the neighborhood of every node. Furthermore, a parameter $n$ is established to regulate the size of the subgraphs. Once the number of sampled nodes reaches $n$, the layer-wise sampling is halted. Since nodes in the sampled neighbors tend to repeat many times, our method could reduce the variance in characterizing the distribution of 1-hop nodes with respect to the source node. The process can be formulated as:

\begin{equation}
    \mathcal{G}_i = (\mathcal{V}_i, \mathcal{E}_i, \boldsymbol{A}_i, \boldsymbol{X}_i) = \text{BFS}(v_i, p, d, n),
\end{equation}
where $\mathcal{G}_i$ is the connected graph obtained, $\mathcal{V}_i$, $\mathcal{E}_i$ are sets of concept nodes and relation links in $\mathcal{G}_i$, $\boldsymbol{A}_i$ represents the adjacent matrix, $\boldsymbol{X}_i$ denotes embeddings of concept nodes, and $v_i$ is the central concept node.

\begin{figure*}[t!]
    \centering
    \setlength{\abovecaptionskip}{0.15cm}
    \includegraphics[scale=0.52]{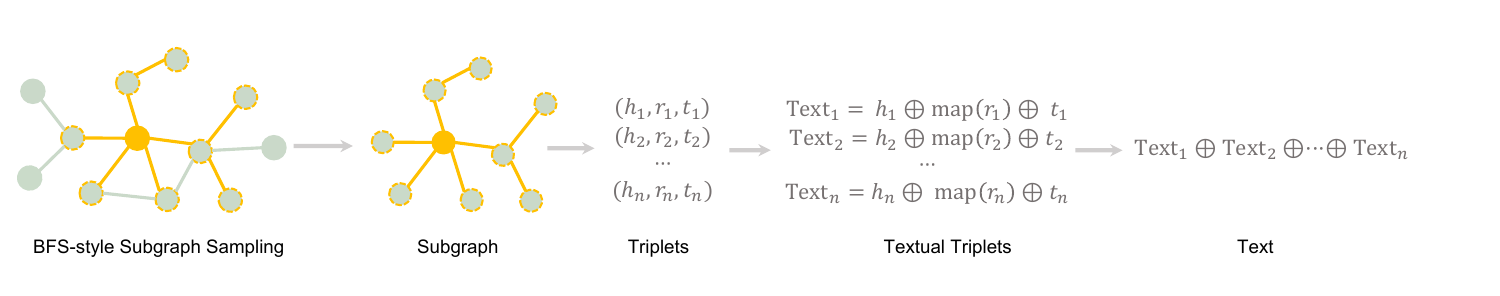}
    \caption{The overview of the construction of graph-text pairs.}
    \label{fig:appendix}
\end{figure*}

After that, it is necessary to textualize the subgraphs to construct synonymous text descriptions. The first step is to convert all relation links into triplet descriptions, which are later combined to compose the final description. Specifically, to transform relation links into sentences, we first map each relation type to a relation template and then concatenate the head concept, relation template, and tail concept as the description of each fact triplet. The textualization process can be denoted as:
\begin{equation}
    s_i = \bigoplus_{e_j \in \mathcal{E}_i} \text{TEXT}(e_j),
\end{equation}
where $s_i$ is the description of the graph $\mathcal{G}_i$ and $\oplus$ denotes the concatenation of sentences. Therefore, the training set can be denoted as $\{(\mathcal{G}_i, s_i) \}_{i=1}^n$. The overview of the construction process is shown in Figure~\ref{fig:appendix}.

\subsubsection{Graph-Text Contrastive Learning}
The graph-text alignment procedure is presented in the left part of Figure~\ref{fig:overview}. First, GNN and PLM are utilized to encode the knowledge subgraphs and natural language descriptions to obtain the corresponding representation, respectively, which can be formulated as:
\begin{equation}
    \begin{aligned}
    \boldsymbol{\tilde{e}_i} &= \text{Pool}_G(\text{GNN}(\mathcal{G}_i)), \\
    \boldsymbol{\tilde{h}_i} &= \text{Pool}_T(\text{PLM}(s_i)),
    \end{aligned}
\end{equation}
where $\boldsymbol{\tilde{e}_i}$  is the average of all nodes' embeddings and $\boldsymbol{\tilde{h}_i}$ is the representation of the [CLS] token. To project $\boldsymbol{\tilde{e}_i}$ and $\boldsymbol{\tilde{h}_i}$ into the same semantic space, two linear projection layers are designed as follows:
\begin{equation}
    \begin{aligned}
    \boldsymbol{e}_i &= \boldsymbol{W}_G \boldsymbol{\tilde{e}_i} + \boldsymbol{b}_G, \\
    \boldsymbol{h}_i &= \boldsymbol{W}_T \boldsymbol{\tilde{h}_i} + \boldsymbol{b}_T,
    \end{aligned}
\end{equation}
where $\boldsymbol{W}_G$, $\boldsymbol{W}_T$ and $\boldsymbol{b}_G$, $\boldsymbol{b}_T$ are the transform matrices and biases of the linear projection layers.

We then employ InfoNCE with in-batch negative sampling to align the representations of two modalities bidirectionally. The graph-to-text contrastive loss can be formulated as:
\begin{equation}
    \mathcal{L}_{\text{G2T}} = -\frac{1}{N}\sum\limits_{i=1}^N\log\frac{\exp(sim(\boldsymbol{e}_i, \boldsymbol{h}_i) / \tau)}{\sum_{j=1}^N \exp(sim(\boldsymbol{e}_i,\boldsymbol{h}_j) / \tau)},
\end{equation}
where $\tau$ is a temperature coefficient and $N$ is the number of instances in a batch. Besides, function $sim(\cdot, \cdot)$ measures the similarity between two representations, which can be calculated by:
\begin{equation}
    sim(\boldsymbol{e_i},\boldsymbol{h_i}) = \frac{\boldsymbol{e_i}^T\boldsymbol{h}_i}{\lVert \boldsymbol{e_i}\rVert \cdot \lVert \boldsymbol{h_i}\rVert}.
\end{equation}

Similarly, we also design a text-to-graph contrastive loss for uniformly aligning into the same semantic space, which is shown as:
\begin{equation}
    \mathcal{L}_{\text{T2G}} = -\frac{1}{N}\sum\limits_{i=1}^N\log\frac{\exp(sim(\boldsymbol{h}_i, \boldsymbol{e}_i) / \tau)}{\sum_{j=1}^N \exp(sim(\boldsymbol{h}_i,\boldsymbol{e}_j) / \tau)}.
\end{equation}

The final contrastive loss $\mathcal{L}_{GT}$ is defined as the average of $\mathcal{L}_{G2T}$ and $\mathcal{L}_{T2G}$:
\begin{equation}
    \mathcal{L}_{GT} = \frac{1}{2}(\mathcal{L}_{G2T} + \mathcal{L}_{T2G}).
\end{equation}

\paragraph{Remarks} (1) To avoid the loss of inherent knowledge caused by over-fitting of the PLM, only the GNN and the linear projection layers are trainable during actual implementation. From another perspective, we essentially distill the semantic information from the PLM into the GNN, enabling the graph representations encoded by the GNN to encompass both structural and textual information. (2) As our ultimate goal is to obtain subgraph representations, we employ graph-level contrastive learning to align subgraph embeddings with text embeddings, yielding promising results. We also attempt other granular alignment signals, such as aligning entity node representations with text representations or applying the Masked Language Model (MLM) to text based on subgraph representations. However, with our current computational resources, these methods are unable to converge effectively. Through our contrastive learning, the model is able to rapidly converge and achieve promising performance.

\subsection{Subgraph Retrieval Module}
\label{sec:retrieval}
In this section, we initially present the establishment of the subgraph vector database. After that, to better accommodate the alignment process, we propose the query enhancement. Finally, we will outline the subgraph retrieval procedure. 

\subsubsection{Database Construction}
Previous methodologies primarily adopt an extracting-and-modeling paradigm for knowledge subgraph retrieval. However, as the cornerstone of subsequent work, the retrieval of high-quality subgraphs proves to be challenging. Consequently, we suggest transforming the knowledge graph into a subgraph vector database, thereby transitioning the focus towards retrieving pertinent subgraphs. Leveraging the analogy between BFS and the message-passing mechanism, we adopt a BFS-style subgraph sampling strategy to construct subgraphs, instead of DFS or Random Walk. On the one hand, each subgraph contains complete neighbor information for at least one node, and on the other hand, each node appears in at least one subgraph. Therefore, the information supplied for our retrieval is complete, and each subgraph vector holds fine-grained knowledge regarding the central node. Specifically, we first apply the same method as in Section~\ref{sec:alignment} to produce the graph embedding $\boldsymbol{e}_i$ and the text embedding $\boldsymbol{h}_i$. Then we add them up as the subgraph vector $\boldsymbol{g}_i \in \mathbb{R}^d$:
\begin{equation}
    \boldsymbol{g}_i = \frac{1}{2}(\frac{\lVert \boldsymbol{h}_i \rVert}{\lVert \boldsymbol{e}_i \rVert}\boldsymbol{e}_i + \boldsymbol{h}_i),
\end{equation}
where the regularization coefficient $\frac{\lVert \boldsymbol{h}_i \rVert}{\lVert \boldsymbol{e}_i \rVert}$ maintains the consistency between the norm of the subgraph vectors and the norm of the text representations, preventing the prediction from relying predominantly on the features with larger norms. Finally, a subgraph vector database $\boldsymbol{G} = \{\boldsymbol{g}_i\}_{i=1}^{\lvert G \rvert}$ is constructed with all subgraph vectors. Note that we only need to generate subgraph vectors once before performing downstream tasks, which saves computational resources.

\subsubsection{Query Enhancement}
Given a problem, we need to find the relevant subgraph vectors. An intuitive method is to apply the embedding of the question-answer pair as a query. However, there is a certain difference between such textual query and the pre-trained corpus of the aligned encoder, as the latter is constructed through triplet concatenation, which makes it difficult to ensure the quality of text encoding and reduces the accuracy of the retrieval process. Therefore, we propose to enhance the query by retrieving question-related triplets in the knowledge graph and concatenating them after the Q-A pairs. Specifically, given a pair of question-answers $(q, c_i)$, we first apply entity linking to find all entities $E_q = \{e_q^{(1)},e_q^{(2)},\dots,e_q^{(n_q)}\}$, $E_{c_i} = \{e_{c_i}^{(1)},e_{c_i}^{(2)},\dots,e_{c_i}^{(n_{c_i})}\}$ appearing in question $q$ and choice $c$, respectively. Then, we find all triplets in the CSKG containing the entities in $E_q$ and $E_{c_i}$, which can be formulated as $T = \{(e^*, r, e), (e, r, e^*)|e \in E_q \cup E_{c_i}\}$. All fact triplets in $T$ are serialized to natural language sentences and a pre-trained dense retriever is adopted to find the most relevant ones. We concatenate the fact triplets retrieved together, along with the questions and options: $s_i = q \oplus c_i \oplus \text{text}_1 \oplus \text{text}_2 \oplus \dots \oplus \text{text}_K$. The aligned PLM is then utilized to encode $s_i$ to $\boldsymbol{t_i} = \text{PLM}(s_i) \in \mathbb{R}^d$.

\subsubsection{Subgraph Retrieval}
After that, we employ the embedding of Q-A pairs concatenated with factual triples to retrieve the relevant subgraph vectors from the subgraph vector database. As the embedding space of two modalities has been aligned, the cosine similarity of $\boldsymbol{t}_i$ with each subgraph vector $\boldsymbol{g}_i$ in $\boldsymbol{G}$ is competent for the retrieval. We recall the top $k$ subgraph vectors with the highest similarities, which is denoted as $\boldsymbol{G}_{q, c_i} \in \mathbb{R}^{k \times d}$. 

\subsection{Prediction}
\label{sec:prediction}
We combine all the knowledge retrieved to make the final predictions. We first integrate the retrieved subgraph vectors through multi-head attention with $\boldsymbol{t}_i$ as the query, which can be formulated as:
\begin{equation}
    \begin{aligned}
        \boldsymbol{\alpha}_i^{(h)} &= \frac{(\boldsymbol{t}_i\boldsymbol{W}_Q^{(h)})(\boldsymbol{G}_{q,c_i}\boldsymbol{W}_K^{(h)})^T}{\sqrt d}, \\
        \boldsymbol{r}_i^{(h)} &= \text{Softmax}(\boldsymbol{\alpha}_i^{(h)})(\boldsymbol{G}_{q,c_i}\boldsymbol{W}_V^{(h)}), \\
        \boldsymbol{r}_i &= \text{Concat}(\boldsymbol{r}_i^{(1)},\boldsymbol{r}_i^{(2)},\dots,\boldsymbol{r}_i^{(H)})\boldsymbol{W}_O,
    \end{aligned}
\end{equation}
where $\boldsymbol{W}_Q^{(h)},\boldsymbol{W}_K^{(h)},\boldsymbol{W}_V^{(h)} \in \mathbb{R}^{d \times d}$ are projection matrices under head $h$.

Subsequently, $\boldsymbol{r}_i$ and $\boldsymbol{t}_i$ are added and fed into a linear layer to predict the score of option $c_i$:
\begin{equation}
    \hat{p}_i = \boldsymbol{W_1}^T(\boldsymbol{t}_i + \boldsymbol{r}_i) + b_1.
\end{equation}

Since some questions are expected to be answered based solely on the question context, we also encode the Q-A pair $(q, c_i)$ to infer directly:
\begin{equation}
    \begin{aligned}
    \boldsymbol{v_i} &= \text{PLM}(q, c_i), \\
    \tilde{p}_i &= \boldsymbol{W_2}^T \boldsymbol{v}_i + b_2.
    \end{aligned}
\end{equation}
The two scores are weighted and summed to yield the final score:
\begin{equation}
    p_i = \lambda \hat{p}_i + (1 - \lambda)\tilde{p}_i,
\end{equation}
where $\lambda$ is the hyper-parameter for the balance.

During the training phase, we employ the softmax function to normalize the score for each choice and optimize the model by cross-entropy loss. For inference, we determine the prediction by selecting the choice with the highest score.

\section{Experiments}
\subsection{Datasets}
We conduct experiments to evaluate our method on five CSQA datasets, which are shown in Table~\ref{tab:dataset}:

$\bullet$ \textbf{CommonsenseQA}~\cite{ref:csqa} is a 5-way multiple-choice QA dataset, which is created based on ConceptNet~\cite{ref:cpnet}. Due to the dual split of CommonsenseQA: the official split~\cite{ref:csqa} and the in-house (IH) split~\cite{ref:kagnet}, we report the results for both settings. For the official split, the ground truth of the test set is not publicly available, so we submit our model's predictions to the official leaderboard\footnote{https://www.tau-nlp.sites.tau.ac.il/csqa-leaderboard} to evaluate the test accuracy.

$\bullet$ \textbf{OpenBookQA}~\cite{ref:obqa} is a 4-choice dataset about elementary science questions to evaluate the science commonsense knowledge. We also submit the predictions of the test set to the official leaderboard\footnote{https://leaderboard.allenai.org/open\_book\_qa/submissions/public}.

$\bullet$ \textbf{SocialIQA}~\cite{ref:socialiqa} is a 3-choice dataset to evaluate the understanding of commonsense social knowledge. Due to the unavailability of the test set, consistent with prior works~\cite{ref:siqa}, we report the accuracy of the development set.

$\bullet$ \textbf{PIQA}~\cite{ref:piqa} is a 2-choice QA dataset regarding physical commonsense. Since the test set is hidden, evaluations are conducted on the development set.

$\bullet$ \textbf{RiddleSenseQA}~\cite{ref:riddlesense} is a 5-choice QA dataset about commonsense riddles. Because the test set is not released, we only report the validation accuracy.
\begin{table}[t!]
    \caption{Statistics of the datasets. '-' denotes the unavailable dataset split.}
    \centering
    \resizebox{0.6\linewidth}{!} {
    \begin{tabular}{l|rrr}
    \toprule
    \textbf{Task} & \textbf{Train} & \textbf{Dev} & \textbf{Test} \\ 
    \midrule
    CommonsenseQA official split & 9,741 & 1,221 & 1,140 \\
    CommonsenseQA in-house split & 8,500 & 1,221 & 1,241 \\
    OpenBookQA & 4,957 & 500 & 500 \\
    SocialIQA & 33,410 & 1,954 & - \\
    PIQA & 16,113 & 1,838 & - \\
    RiddleSenseQA & 3,510 & 1,021 & - \\
    \bottomrule
    \end{tabular}
    }
    \label{tab:dataset}
\end{table}

\subsection{Baselines}
We compare with the mainstream RoBERTa-Large + GNN methods, including RN~\cite{ref:rn}, RGCN~\cite{ref:rgcn}, GconAttn~\cite{ref:gconattn}, MHGRN~\cite{ref:mhgrn}, QA-GNN~\cite{ref:qagnn}, DGRN~\cite{ref:dgrn}, GreaseLM~\cite{ref:greaselm}, JointLK~\cite{ref:jointlk}, GSC~\cite{ref:gsc}, SAFE~\cite{ref:safe}, DRAGON~\cite{ref:dragon}, HamQA~\cite{ref:hamqa}, and DHLK~\cite{ref:dhlk}. Among them, DRAGON introduces BookCorpus to joint-train GNN and PLM, and DHLK additionally retrieves paraphrases of key entities in WordNet and Wiktionary.

\subsection{Implementation Details}
According to the previous works, we use RoBERTa-Large~\cite{ref:roberta} as the text encoder and use GraphGPS~\cite{ref:gps} for the graph encoder. We also test AristoRoBERTa~\cite{ref:aristo} for OpenBookQA. We use ConceptNet~\cite{ref:cpnet}, including 799,273 nodes and 2,487,003 edges, as the commonsense knowledge graph. For each node, we treat it as the center and employ BFS to obtain the subgraph, which is then translated into the corresponding natural language description.

In the graph-text alignment phase, we randomly sample 64,000 graph-text pairs to train, and sample 16,000 pairs to evaluate two encoders. We fix the learning rate to 1$e$-3, the number of GNN layers to $2$, and the dimensions of all embeddings to $1024$. During the fine-tuning stage, we set the number of fact triplets to $10$, tune the number of retrieved subgraph vectors $k$ in $\{10, 30, 50, 70, 100\}$, the batch size in $\{4, 8, 16\}$, the balance coefficient $\lambda$ from $0.1$ to $1.0$, and the learning rate in \{2$e$-5, 1$e$-5, 5$e$-6, 2$e$-6, 1$e$-6\}. The parameters of the model are optimized by RAdam. We train 30 epochs until the performance does not
improve on the development sets for 3 consecutive epochs. We use the default parameter settings as their original implementations for the baseline methods. We conduct all experiments on NVIDIA A100-40GB GPUs.

\subsection{Main Results}
Following previous works~\cite{ref:qagnn,ref:gsc,ref:safe,ref:mvp}, we compare our method with different baselines on CommonsenseQA and OpenBookQA as main results, which are shown in Table~\ref{tab:main}. The best and runner-up results in each column are highlighted in bold and underlined, respectively.
\begin{table}[t!]
  \centering
  \caption{Evaluation on CommonsenseQA (in-house split) and OpenBookQA. We use RoBERTa-Large as the text encoder in CommonsenseQA, and use RoBERTA-Large and AristoRoBERTa in OpenBookQA. Methods with AristoRoBERTa use the textual evidence by~\cite{ref:aristo} as an additional input to the QA context. The baselines incorporating extra corpus are marked with $^*$.}
  \resizebox{0.9\linewidth}{!} {
    \begin{tabular}{l|cc|cc}
    \toprule
    \multicolumn{1}{l|}{\multirow{2}[3]{*}{Methods}} & \multicolumn{2}{c|}{CommonsenseQA} & \multicolumn{2}{c}{OpenBookQA} \\
\cmidrule{2-5}          & IHdev-Acc (\%) & IHtest-Acc (\%) & RoBERTa-Large (\%) & AristoRoBERTa (\%) \\
    \midrule
    Fine-tuned LMs & 73.07 (±0.45) & 68.69 (±0.56) & 64.80 (±2.37) & 78.40 (±1.64) \\
    \midrule
    + RN  & 74.57 (±0.91) & 69.08 (±0.21) & 65.20 (±1.18) & 75.35 (±1.39) \\
    + RGCN & 72.69 (±0.19) & 68.41 (±0.66) & 62.45 (±1.57)  & 74.60 (±2.53) \\
    + GconAttn & 72.61 (±0.39) & 68.59 (±0.96) & 64.75 (±1.48) & 71.80 (±1.21) \\
    + MHGRN & 74.45 (±0.10) & 71.11 (±0.81) & 66.85 (±1.19) & 80.60 \\
    + QA-GNN & 76.54 (±0.21) & 73.41 (±0.92) & 67.80 (±2.75) & 82.77 (±1.56) \\
    + DGRN & 78.20  & 74.00    & 69.60  & 84.10 \\
    + GreaseLM & 78.50 (±0.50) & 74.20 (±0.40) & 68.80 (±1.75) & 84.80 \\
    + JointLK & 77.88 (±0.25) & 74.43 (±0.83) & 70.34 (±0.75) & 84.92 (±1.07) \\
    + GSC & 79.11 (±0.22) & 74.48 (±0.41) & 70.33 (±0.81) & 86.67 (±0.46) \\
    + SAFE & 76.93 (±0.37) & 74.03 (±0.43) & 69.20  & \underline{87.13} \\
    + HamQA & 76.88 & 73.91 & 71.12 & 84.59 \\
    + DRAGON$^*$ & -     & \textbf{76.00}    & 72.00    & - \\
    + DRAGON (w/o MLM)$^*$ & -     & 73.80  & 66.40  & - \\
    + DHLK$^*$ & \underline{79.39} (±0.24) & 74.68 (±0.26) & \underline{72.20} (±0.40) & 86.00 (±0.79) \\
    \midrule
    + SEPTA~(\textbf{Ours})    & \textbf{79.61} (±0.17) & \underline{74.78} (±0.23) & \textbf{72.33} (±0.35) & \textbf{87.37} (±0.51) \\
    \bottomrule
    \end{tabular}%
    }
  \label{tab:main}%
\end{table}%

From the results, we can observe: (1) Our method can contribute performance gains to LMs, which improves 6.54\% and 6.09\% on IHdev and IHtest of CommonsenseQA compared to fine-tuned RoBERTa. (2) SEPTA outperforms all baselines without additional corpus on both datasets. For example, compared to the GSC method, our method improves by 2.00\% and 0.70\% on OpenBookQA using RoBERTa and AristoRoBERTa, respectively. (3) Compared to baselines incorporating additional corpus, our method also achieves comparable performance. Specifically, we surpass DHLK on both datasets and DRAGON on OpenBookQA and slightly lag behind DRAGON on CommonsenseQA. It should be noted that the DRAGON undergoes MLM training on the BookCorpus dataset and requires training on 8$\times$A100 GPUs for a week~\cite{ref:dragon,ref:dhlk}. By eliminating the MLM, our SEPTA model demonstrates a definitive enhancement.

In Table~\ref{tab:leaderboard}, we evaluate SEPTA on the official CommonsenseQA and OpenBookQA leaderboards (as of March 22, 2024). Our method achieves results surpassing all baselines based on the same PLM and exhibits comparative performance compared with methods with larger-scale parameters (e.g., UnifiedQA).
\begin{table}[t!]
\caption{Performance comparison on CommonsenseQA (left) and OpenBookQA (right) official leaderboard.}
\begin{minipage}[c]{0.48\textwidth}
\centering
\resizebox{\linewidth}{!} {
\begin{tabular}{lc}
\toprule
\textbf{Methods} & \textbf{Test-Acc (\%)}  \\
\midrule
RoBERTa~\cite{ref:roberta}& 72.1 \\
RoBERTa+FreeLB  & 72.2 \\
RoBERTa+HyKAS   & 73.2 \\
RoBERTa+KE   & 73.3 \\
RoBERTa+KEDGN      & 74.4 \\
RoBERTa+MHGRN~\cite{ref:mhgrn}     & 75.4 \\
RoBERTa+QA-GNN~\cite{ref:qagnn}       & 76.1 \\
RoBERTa+GSC~\cite{ref:gsc}   & 76.2 \\
Albert   & 73.5 \\
ALBERT+Path Generator~\cite{ref:pg}        & 75.6 \\
ALBERT+HGN~\cite{ref:mhgrn}       & 77.3 \\
UnifiedQA (11B)~\cite{ref:unifiedqa}      & \textbf{79.1} \\
\midrule
RoBERTa+SEPTA~(\textbf{Ours}) & 76.6 \\
\bottomrule
\end{tabular}
}
\end{minipage}
\begin{minipage}[c]{0.52\textwidth}
\centering
\resizebox{\linewidth}{!} {
\begin{tabular}{lc}
\toprule
\textbf{Methods} & \textbf{Test-Acc (\%)}  \\
\midrule
Careful Selection~\cite{ref:selection}& 72.0 \\
AristoRoBERTa~\cite{ref:aristo}       & 77.8 \\
KF+SIR               & 80.0 \\
AristoRoBERTa+PG~\cite{ref:pg}        & 80.2 \\
AristoRoBERTa+MHGRN~\cite{ref:mhgrn}  & 80.6 \\
AristoRoBERTa+QA-GNN~\cite{ref:qagnn}       & 82.8 \\
AristoRoBERTa+GreaseLM~\cite{ref:greaselm}  & 84.8 \\
AristoRoBERTa+GSC~\cite{ref:gsc}  & 87.4 \\
AristoRoBERTa+MVP-Tuning~\cite{ref:mvp}  & 87.6 \\
ALBERT + KB                        & 81.0 \\
T5               & 83.2 \\
UnifiedQA (11B)~\cite{ref:unifiedqa}      & 87.2 \\
\midrule
AristoRoBERTa+SEPTA~(\textbf{Ours}) & \textbf{87.8} \\
\bottomrule
\end{tabular}
}
\end{minipage}
\label{tab:leaderboard}
\end{table}

To comprehensively evaluate the efficiency of SEPTA, we extend our comparative analysis to other commonsense reasoning datasets originating from diverse domains or tasks. As shown in Table~\ref{tab:other_qa}, our SEPTA consistently achieves superior performance. This observation underscores the overall effectiveness of SEPTA in addressing various commonsense reasoning datasets or tasks, demonstrating a unified methodology.
\begin{table}[t!]
\centering
\caption{Performance comparison on SocialIQA, PIQA, and RiddleSenseQA.}
\resizebox{0.55\linewidth}{!} {
\begin{tabular}{lcccc}
    \toprule  
    \textbf{Methods}& \textbf{SocialIQA} & \textbf{PIQA} & \textbf{RiddleSenseQA}\\
    \midrule  
    RoBERTa-Large  & 78.25 & 77.53 & 60.72\\
    \midrule
    + {GconAttn} & 78.86 & 78.24 & 61.77 \\
    + {RN} & 78.45 & 76.88 &62.17 \\
    + {MHGRN} & 78.11 & 77.15 & 63.27\\
    + {QA-GNN} & 78.10 & 78.24 & 63.39 \\
    + {GreaseLM} & 77.89 & 78.02 & 63.88 \\
    + {GSC} & 78.61 &  78.40 & 64.07 \\
    + {SAFE} & 78.86 & 79.43 & 63.78 \\
    \midrule
    + {SEPTA}~(\textbf{Ours}) &  \textbf{79.21}  & \textbf{80.85} &  \textbf{67.62} \\
    \bottomrule 
\end{tabular}
}
\label{tab:other_qa}
\end{table}

\subsection{Ablation Study}
We conduct an ablation study on CommonsenseQA and OpenBookQA to explore the effectiveness of each component of SEPTA. We remove the alignment process (w/o alignment), retrieved subgraph vectors (w/o subgraph), fact triplets (w/o triplets), and scores predicted based on Q-A pairs (i.e. set $\lambda=1.0$), respectively.
\begin{table}[t!]
\centering
\caption{Ablation study on CommonsenseQA (IHtest) and OpenBookQA datasets. The values in parentheses denote the extent of performance decline.}
\resizebox{0.6\linewidth}{!} {
\begin{tabular}{lcc}
    \toprule  
    \textbf{Ablation}& \textbf{CommonsenseQA} & \textbf{OpenBookQA}\\
    \midrule
    SEPTA & 74.78 & 72.33 \\
    \midrule
    w/o alignment & 69.83 (-4.95) & 67.20 (-5.13) \\
    w/o subgraph  & 72.34 (-2.44) & 70.23 (-2.10) \\
    w/o triplets  & 71.25 (-3.53) & 69.67 (-2.66) \\
    $\lambda=1.0$ & 74.13 (-0.65) & 70.47 (-1.86)  \\
    \bottomrule 
\end{tabular}
}
\label{tab:ablation}
\end{table}

As shown in Table~\ref{tab:ablation}, four components are all crucial for SEPTA, and removing any part will result in a decrease in performance. Specifically, the performance drops the most significantly when we remove the graph-text alignment. This is because if the representations of graphs and texts are not semantically aligned, then during the knowledge retrieval stage, the retrieved subgraph vectors may be irrelevant and situated in different latent spaces from the textual information. Moreover, removing either fact triplets or subgraph vectors will affect the performance. On one hand, they represent different aspects of information, with the former providing more specific knowledge and the latter describing more comprehensive relationships between entities. On the other hand, fact triplets also play an auxiliary role in retrieving relevant subgraph vectors. Furthermore, only using knowledge-enhanced representations for predictions (i.e. $\lambda=1.0$) cannot achieve optimal results. This is because some questions do not require additional knowledge, or relevant information cannot be found in CSKGs, which may instead become interference.

\subsection{Low-Resource Setting}
To evaluate the robustness of SEPTA, we conduct extensive experiments in low-resource settings, with different proportions of training data, including 5\%, 10\%, 20\%, 50\%, and 80\%, in CommonsenseQA (IHtest) and OpenBookQA. 

From the results in Table~\ref{tab:few-shot}, we can observe that our SEPTA achieves promising performance in all settings, and it exhibits a trend where the performance improvement relative to other baselines is more significant with fewer training data. This is because we align text representations with graph representations before fine-tuning on downstream tasks, enabling retrieved subgraph vectors to integrate well with text representations even in low-resource settings. In contrast, other baselines make it hard to project subgraph representations and text representations into the same semantic space when training data is limited, resulting in structure embeddings becoming a noise that interferes with PLM reasoning.
\begin{table}[t!]
\centering
\caption{Performance with different proportions of training data.}
\resizebox{0.95\linewidth}{!} {
\begin{tabular}{lcccccccccccc}
    \toprule
        \multirow{2}{*}{\textbf{Methods}}&
        \multicolumn{6}{c}{CommonsenseQA} & \multicolumn{6}{c}{OpenBookQA}\\
        \cmidrule(lr){2-7} \cmidrule(lr){8-13}
        & 5\% & 10\% & 20\% & 50\% & 80\% & 100\% & 5\% & 10\% & 20\% & 50\% & 80\%  & 100\%\\
        \midrule
        RoBERTa-large & 29.66 & 42.84 & 58.47 & 66.13 & 68.47 & 68.69 & 37.00 & 39.4 & 41.47 & 53.07 & 57.93 & 64.8\\
        \midrule
        + RGCN & 24.41 & 43.75 & 59.44 & 66.07 & 68.33 & 68.41 & 38.67 & 37.53 & 43.67 & 56.33 & 63.73 & 62.45\\
        + GconAttn & 21.92 & 49.83 & 60.09 & 66.93 & 69.14 & 68.59 & 38.60 & 36.13 & 43.93 & 50.87 & 57.87 & 64.75\\
        + RN & 23.77 & 34.09 & 59.90 & 65.62 & 67.37 & 69.08 & 33.73 & 35.93 & 41.40 & 49.47 & 59.00 & 65.20\\
        + MHGRN & 29.01 & 32.02 & 50.23 & 68.09 & 70.83 & 71.11 & 38.00 & 36.47 & 39.73 & 55.73 & 55.00 & 66.85\\
        + QA-GNN & 32.95 & 37.77 & 50.15 & 69.33 & 70.99 & 73.41 & 33.53 & 35.07 & 42.40 & 54.53 & 52.47 & 67.80\\
        + GreaseLM & 22.80 & 56.16 & 63.09 & 70.56& 73.41& 74.20& 39.00&39.60&42.20&57.87 & 65.13&68.80\\
        + GSC & 31.02 & 35.07 & 65.83 & 70.94& 73.82& 74.48 & 29.60 & 41.80 & 42.40 & 58.03 & 65.97 & 70.33\\
        + SAFE & 36.45 & 56.51 & 65.16 & 70.72 & 73.22 & 74.03 & 38.80 & 41.20 & 44.93 & 58.33 & 65.60 & 69.20 \\
        \midrule
        + SEPTA(\textbf{Ours}) & \textbf{50.69} & \textbf{62.37} & \textbf{68.09} & \textbf{71.80} & \textbf{74.05} & \textbf{74.78} & \textbf{45.63} & \textbf{54.80} & \textbf{58.10} & \textbf{66.57} & \textbf{68.30} & \textbf{72.33}\\
    \bottomrule
\end{tabular}
}
\label{tab:few-shot}
\end{table}

\subsection{Evaluation with Other GNNs}
To demonstrate the generality of SEPTA, We employ GraphGPS~\cite{ref:gps}, FILM-GNN~\cite{ref:film}, and RGCN~\cite{ref:rgcn} as the graph encoders, respectively. Table~\ref{tab:gnn} illustrates the results on CommonsenseQA and OpenBookQA. From the results, we can observe that different graph encoders achieve competitive results on both datasets, with their performances being relatively close (within a difference of around 0.5\%), which demonstrates the effectiveness and robustness of SEPTA.
\begin{table}[t!]
\centering
\caption{Effect of different graph encoders.}
\resizebox{0.55\linewidth}{!} {
\begin{tabular}{lcc}
    \toprule  
    \textbf{GNN}& \textbf{CommonsenseQA} & \textbf{OpenBookQA}\\
    \midrule
    GraphGPS & 74.78 & 72.33 \\
    FILM-GNN & 74.67 & 72.17 \\
    RGCN & 74.51 & 71.87 \\
    \bottomrule 
\end{tabular}
}
\label{tab:gnn}
\end{table}

\subsection{Hyper-parameter Analysis}
\begin{figure}[t!]
  \centering
  \setlength{\abovecaptionskip}{-0.15cm}
    \subfigure{
    \setcounter{subfigure}{0}
    \subfigure[Effect of $k$.]{
    \includegraphics[width=0.30\linewidth]{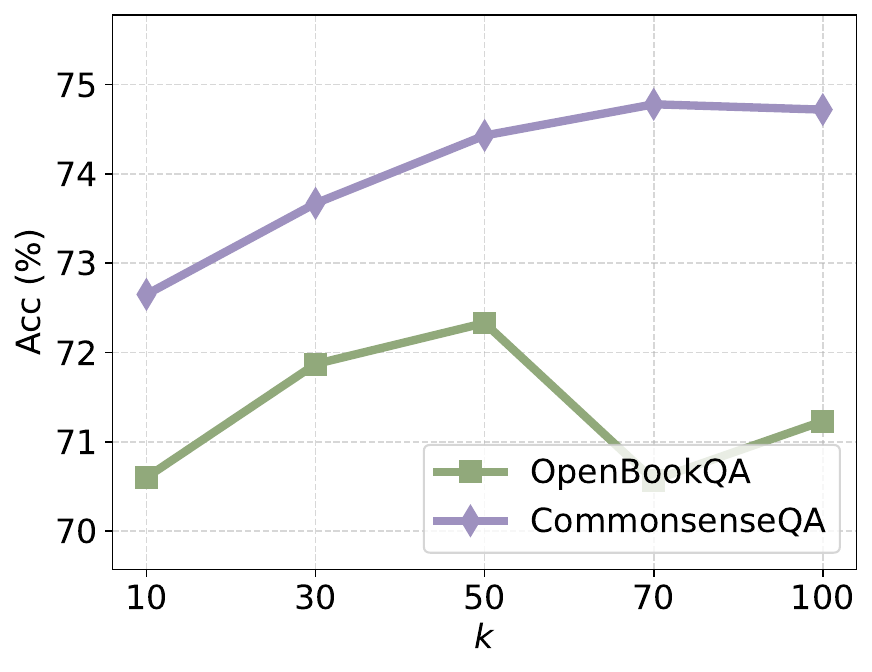}
    }
    \subfigure[Effect of $n$.]{
    \includegraphics[width=0.30\linewidth]{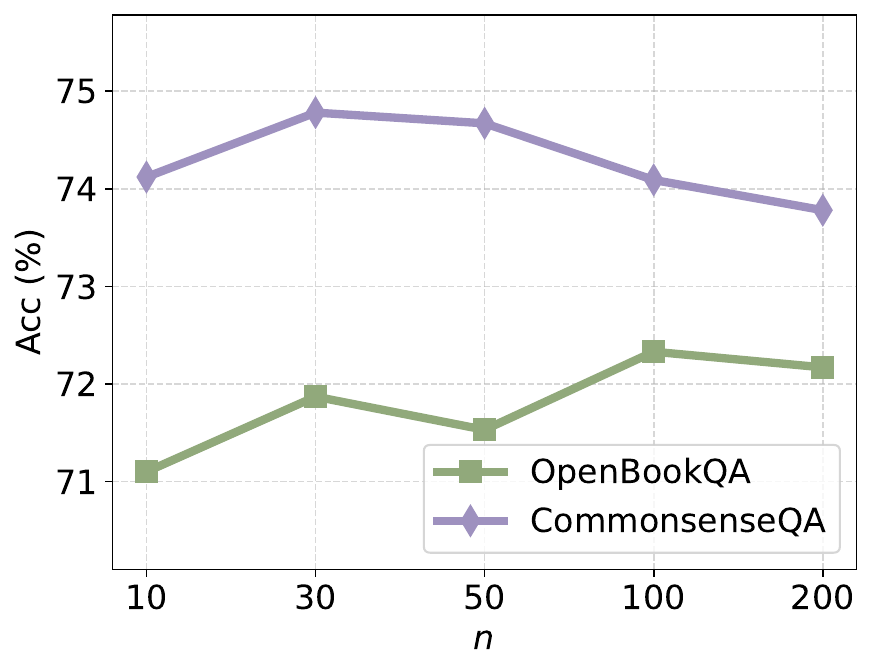}
    }
    \subfigure[Effect of $\lambda$.]{
    \includegraphics[width=0.30\linewidth]{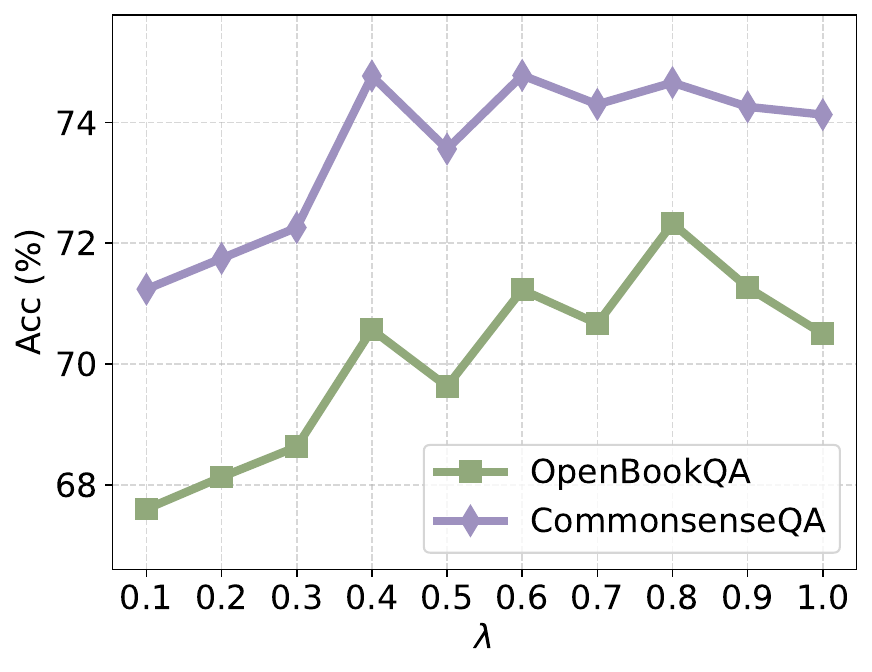}
    }
    }
  \caption{Hyper-parameter analysis.}
  \label{fig:hyperparameter}
\end{figure}

We further conduct in-depth analyses to investigate the impact of hyper-parameters. With other parameters fixed, we compare the effect
of the number of retrieved subgraph vectors $k$, the maximum number of nodes $n$ in each subgraph, and the balance coefficient $\lambda$. The results on CommonsenseQA (IHtest) and OpenBookQA datasets are reported in Figure~\ref{fig:hyperparameter}.

\subsubsection{Effect of subgraph vectors number $k$} From the results, we observe that the accuracy
initially ascends with the increase in the number of retrieved subgraph vectors, achieving its peak
before subsequently declining. This is because fewer subgraph vectors may lose crucial commonsense, while an excess of subgraph vectors could introduce irrelevant information.

\subsubsection{Effect of maximum number of nodes $n$} Based on the results, the performance of the model initially increases with the increment of $n$, reaching a peak, then decreases. It might be attributed to the fact that when $n$ is relatively small, subgraphs are unable to fully encompass the neighbor information of the central nodes, leading to the inability to acquire sufficient relevant knowledge during the subgraph retrieval phase. Conversely, when $n$ is excessively large, each subgraph may contain a significant amount of information irrelevant to the central node, resulting in overall information redundancy.

\subsubsection{Effect of the balance coefficient $\lambda$} $\lambda$ controls the proportion of inference based on the retrieved knowledge. When $\lambda$ is small, the model primarily relies on its own knowledge for inference, which may lead to a lack of relevant information for some questions. However, when $\lambda$ is large, the model heavily depends on retrieved knowledge to derive answers, although many questions need to be resolved according to the question context. Therefore, in terms of results, the accuracy generally increases initially with the increase in $\lambda$ and then decreases.

\section{Ethical Considerations and Limitations}
\subsection{Ethical Considerations}
Our work proposes a novel and effective framework to combine PLMs and external knowledge graphs for commonsense question answering. However, potential issues may arise from the utilization of PLMs and CSKGs. On one hand, PLMs tend to encapsulate certain biases present in the pre-training data. On the other hand, CSKGs may harbor biased concepts stemming from human annotations. To alleviate these biases, the implementation of appropriate screening rules offers a promising approach, e.g., filter biased concepts during the subgraph extraction process from the CSKG. Although a comprehensive analysis of such biases is not included in our work, it is imperative to implement supplementary measures before deploying the system in real-world scenarios.

\subsection{Limitations}
We propose a subgraph retrieval enhanced by a graph-text alignment framework named SEPTA for commonsense question answering. However, there are still limitations that demand resolution. Firstly, the corresponding text generated by rules from knowledge subgraphs still exhibits disparities from natural language. One possible solution is to reorganize the text using LLMs, but the cost is prohibitively high. Therefore, acquiring large-scale, high-quality graph-text pairs remains an ongoing challenge. Secondly, the number of retrieved subgraph vectors is required to tune according to the accuracy of development sets, which is time-consuming. Designing a module to automatically select the number may be a solution worth exploring. Thirdly, due to considerations of fairness in comparison and limited computational resources, we do not employ other PLMs, especially LLMs, as text encoders, which will be considered in our future work.

\section{Conclusion}
We propose an effective framework: subgraph retrieval enhanced by graph-text alignment, named SEPTA, for commonsense question answering. In our method, we reframe the task as a subgraph vector retrieval problem and introduce a graph-text alignment method to enhance retrieval accuracy and facilitate knowledge fusion for prediction. Subsequently, all the structural information retrieved is then combined by a simple attention mechanism to bolster the reasoning capabilities of PLMs. Extensive experiments on five benchmarks demonstrate the effectiveness of SEPTA.

In the future, our work will focus on the following aspects. First, we will explore more efficacious pre-training tasks for semantic alignment. Second, if there are sufficient computational resources, we intend to apply our approach to larger language models. Third, we will try SEPTA on relevant tasks, e.g., node classification and link predictions on text-attributed graphs.

\begin{credits}
\subsubsection{\ackname} This work is supported by Ant Group through Ant Research Intern Program.

\subsubsection{\discintname}
The authors have no competing interests to declare that are relevant to the content of this article.
\end{credits}


\end{document}